\documentclass[10pt,twocolumn,letterpaper]{article}

\usepackage{wacv}              

\usepackage{graphicx}
\usepackage{amsmath}
\usepackage{amssymb}
\usepackage{booktabs}

\usepackage[pagebackref,breaklinks,colorlinks]{hyperref}

\usepackage[capitalize]{cleveref}
\crefname{section}{Sec.}{Secs.}
\Crefname{section}{Section}{Sections}
\Crefname{table}{Table}{Tables}
\crefname{table}{Tab.}{Tabs.}

\def\wacvPaperID{1869} 
\def\confName{WACV}
\def\confYear{2024}

\begin{document}

\title{On quantifying and improving realism of images generated with diffusion}

\author{
Yunzhuo Chen\textsuperscript{1}, Naveed Akhtar\textsuperscript{2}, Nur Al Hasan Haldar\textsuperscript{3}, Ajmal Mian\textsuperscript{4}\\
\small The University of Western Australia, Perth, Australia \\
\textsuperscript{1}\texttt{\small yunzhuo.chen@research.uwa.edu.au},
\textsuperscript{2}\texttt{\small naveed.akhtar@uwa.edu.au}, \\
\textsuperscript{3}\texttt{\small nur.haldar@uwa.edu.au},
\textsuperscript{4}\texttt{\small ajmal.mian@uwa.edu.au}
}
\maketitle

\maketitle

\begin{abstract}
\vspace{-2mm}
   Recent advances in diffusion models have led to a quantum leap in the quality of generative visual content. However, quantification of  realism of the  content is still challenging. Existing evaluation metrics, such as Inception Score and Fréchet inception distance, fall short on benchmarking diffusion models due to the versatility of the generated images. Moreover, they are not designed to quantify  realism of an individual image. This restricts their application in forensic image analysis, which is becoming increasingly important in the emerging era of generative models. To address that, we first propose a metric, called Image Realism Score (IRS), computed from five statistical measures of a given image. This non-learning based metric not only efficiently quantifies realism of the generated images, it is readily usable as a measure to classify a given image as real or fake. 
   We experimentally establish the model- and data-agnostic nature of the proposed IRS by successfully detecting fake images generated by Stable Diffusion Model (SDM), Dalle2, Midjourney and BigGAN.  
   We further leverage this attribute of our metric to minimize an IRS-augmented generative loss of SDM, and demonstrate a convenient yet considerable quality improvement of the SDM-generated content with our modification. Our efforts have also led to Gen-100 dataset, which provides 1,000 samples for 100 classes generated by four high-quality models.  We will release the dataset and code.
\end{abstract}

\vspace{-6mm}
\section{Introduction}
\vspace{-1mm}

Generative models, including Variational Autoencoders (VAEs)~\cite{doersch2016tutorial, oussidi2018deep}, Energy-Based Models (EBM)~\cite{lecun2006tutorial, ngiam2011learning}, Generative Adversarial Network (GANs)~\cite{creswell2018generative, wang2017generative} and  normalizing  flow~\cite{rezende2015variational} have historically attracted significant attention from the research community, only to be eventually surpassed by diffusion models~\cite{ho2020denoising, cao2022survey}. Diffusion models have recently provided a quantum leap to the generative visual content quality~\cite{cao2022survey}. 
Moreover, they are claimed to also overcome challenges such as matching posterior distributions in VAEs, managing unpredictability in GAN objectives, high computational demands of Markov Chain Monte Carlo \cite{geyer1992practical} techniques in EBMs, and network limitations of the 
normalizing flows. Naturally, we can expect an even higher popularity of the diffusion models in generative computer vision in the future. 

\begin{figure}[t]
  \centering
\includegraphics[width=\columnwidth]{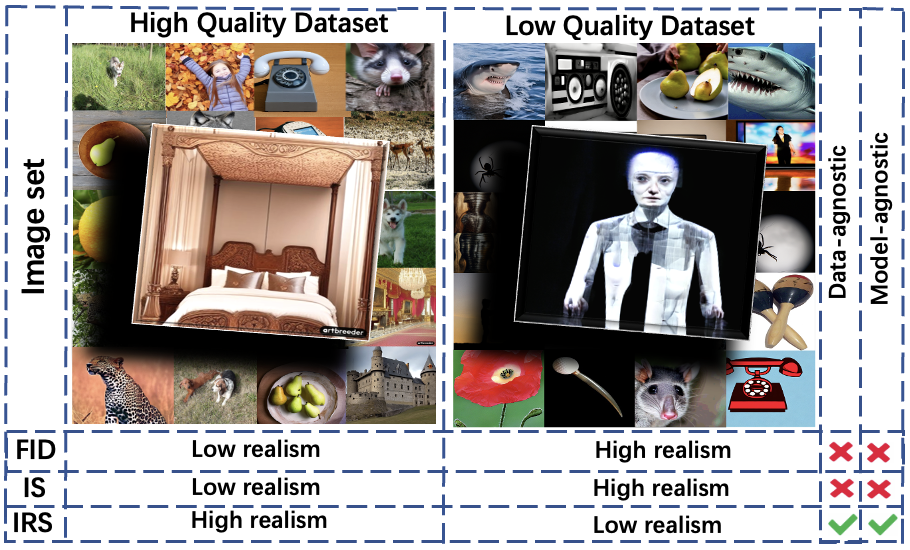}
   \vspace{-7mm}
\caption{
   We manually group images into high and low quality sets and measure their realism scores. FID~\cite{heusel2017gans} and IS~\cite{salimans2016improved} are model/data-dependent metrics and produce counter-intuitive results on these sets. The proposed Image Realism Score (IRS) is sample-specific and performs intuitive and reliable discrimination between the low and high quality samples. Unlike FID and IS, the proposed IRS can also be used as a loss to improve the generative ability of a model.}    
   \label{intro}
   \vspace{-3mm}
\end{figure}

\begin{figure*}[t]
  \centering
  \includegraphics[width=1\linewidth]{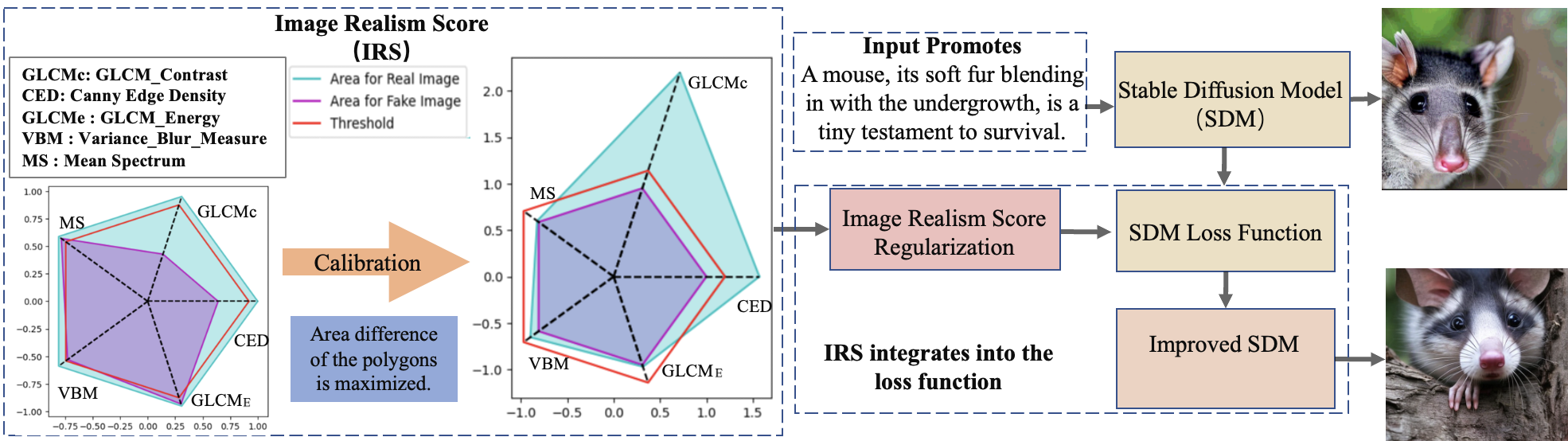}
\vspace{-3mm}
   \caption{Overall framework of the proposed IRS. The left side illustrates how IRS quantifies realism in an image through pentagon area; larger area means higher realism. IRS maximizes the area difference between real and generated images through a special arrangement of five statistical measures followed by a calibration process. The right side shows that IRS is used to modify the loss function to improve  realism in the generative model outputs, taking SDM as an example.}
   \label{Frame}
   \vspace{-2mm}
\end{figure*}

Owing to the importance of generative modeling in vision, a number of metrics have been proposed to evaluate the abilities of generative models~\cite{binkowski2018demystifying,heusel2017gans,salimans2016improved,zhou2019hype}.
However, due to the high quality and versatility of the content generated by diffusion models, these metrics are now falling short on providing meaningful evaluation of the diffusion model generated content. Not to mention, these metrics have widely known intrinsic weaknesses. 
For instance, the popular Fréchet Inception Distance (FID)~\cite{heusel2017gans} is known to suffer from notable bias~\cite{binkowski2018demystifying}. The Inception Score (IS)~\cite{salimans2016improved} is also known to be often suboptimal~\cite{barratt2018note}. Moreover, both metrics are dataset- or model-dependent in the sense that they rely on a reference dataset or a model to compute their scores. For example, IS~\cite{salimans2016improved} uses an ImageNet~\cite{russakovsky2015imagenet} trained Inception model~\cite{szegedy2017inception} to provide a meaningful evaluation score. This is particularly problematic  for forensics of generative content where normally a single sample is available and the task is to adjudge its authenticity in a model- and data-agnostic manner. 


Nowadays, we are witnessing many new generative diffusion models surfacing on a daily basis, each with better content quality and flexibility than the previous ones~\cite{cao2022survey}. With public access to these models, this can lead to serious societal problems if the generated content are used with a negative intent~\cite{roettgers2018porn}. The inability of the current evaluation metrics for generative modeling to verify the authenticity of individual (generated) images is an obvious shortcoming that needs to be addressed for the forensic treatment of the content~\cite{chong2020effectively}. This work fills-in the gap by proposing an Image Realism Score (IRS) metric that is more suited to benchmark the content quality of modern diffusion models. 


The proposed IRS is a `non-learning' based metric which allows it to be dataset- and model-agnostic, while computing intuitive scores according to image quality, see Fig.~\ref{intro}. The proposed metric relies on well-established concepts in image processing, including   Canny Edge Density~\cite{tahmid2017density}, GLCM Contrast~\cite{sebastian2012gray}, GLCM Energy~\cite{sebastian2012gray}, Variance of Laplacian~\cite{he2005laplacian}, and Mean Spectrum~\cite{frank2020leveraging} to determine the realism of an image. Leveraging the sample-specific nature of the statistics used by IRS, we can delineate between natural and fake content easily with our metric. To facilitate further efforts towards the quantification of realism in diffusion-generated content, we also introduce a 100-category dataset in this work, called Gen-100. Each category in this dataset contains 1,000 images generated with ChatGPT prompts using  Stable Diffusion Model (SDM)\cite{rombach2022high}, Dalle2\cite{daras2022discovering}, Midjourney\cite{rescher1983mid} and BigGAN\cite{chang2020tinygan}.
Our dataset is mainly used to validate the model-agnostic and data-agnostic nature of IRS. We eventually exploit the same  property of IRS to minimize the generative loss of SDMs, which allows us to considerably improve the quality of SDM-generated content. To summarize, this paper makes the following three main contributions.
\begin{itemize}
    \item It introduces Image Realism Score (IRS), a first-of-its-kind non-learning based sample specific metric to quantify realism of an image for differentiating natural images from those generated by generative models. 
\vspace{-2mm}
    \item It leverages IRS to benchmark realism of popular generative models and establishes that the  metric is well-suited to forensic analysis by employing it for fake image detection. In the process, it also proposes Gen-100 dataset that contains 1,000 images of 100 classes generated by four models. 
\vspace{-2mm}
    \item By regulating the training loss of Stable Diffusion Model (SDM)~\cite{rombach2022high} to further minimize the proposed metric score, it demonstrates a considerable improvement in the quality of images generated by SDM. 
\end{itemize}

\vspace{-3mm}
\section{Related Work}
\vspace{-1mm}
In the last decade, high-quality sample generation of Generative Adversarial Networks (GANs) \cite{brock2018large} has enabled deep generative models to  receive widespread attention. Nevertheless, with the emergence of diffusion models \cite{binkowski2018demystifying, heusel2017gans,salimans2016improved,zhou2019hype,rombach2022high} GANs are no longer the dominant force in this field. Diffusion models have gained rapid popularity in recent years owing to their stability and superior generation quality. They address some of the common challenges associated with GANs, such as mode collapse, the overhead of adversarial learning, and convergence failure~\cite{alqahtani2019analysis}. The training strategy of diffusion models involves systematically corrupting the training data by gradually adding Gaussian noise, followed by learning to retrieve the original data from the noisy version~\cite{cao2022survey}. Additionally, since their training approach makes small changes to the original data and then corrects those changes, they manage to learn a data distribution where samples closely follow the original data, providing a strong sense of realism to the generated samples. These strengths of diffusion models have led to their significant achievements in the field of image generation technologies \cite{alqahtani2019analysis, ramesh2022hierarchical}. Moreover, diffusion models have been widely applied in various domains, including image denoising~\cite{nichol2021improved} and repair~\cite{esser2021imagebart}, image super-resolution~\cite{li2022srdiff, batzolis2021conditional}, and text-to-image generation~\cite{rombach2022text,saharia2022photorealistic}.

The diffusion model training process consists of two steps. First, there is a predefined forward process that transforms the data distribution into a Gaussian distribution. The second, the corresponding reverse process, employs a trained neural network to simulate regular or random stepwise reversal of the forward process. Diffusion modeling provides a more stable training target and higher generative quality compared to VAEs, EBMs, and normalizing flows~\cite{doersch2016tutorial,oussidi2018deep,creswell2018generative,wang2017generative}. However, because the prior distribution is iteratively transformed into a complex data distribution, a significant number of function evaluations are required in the reverse process. Consequently, diffusion models inherently suffer from a more time-consuming sampling process. Researchers have proposed various solutions, such as introducing new forward processes to stabilize sampling~\cite{hoogeboom2021autoregressive,de2021diffusion}, and recent studies have addressed this issue by implementing dimensionality reduction~\cite{giannone2022few,hoogeboom2021argmax}.

Although several evaluation metrics are proposed to measure the performance of deep generative models, there is no globally accepted agreement on the best metrics for generative models. Currently, popular metrics include the Inception Score (IS), the Frechet Inception Distance (FID), Maximum Mean Discrepancy (MMD)\cite{fortet1953convergence} and Activation maximization score\cite{zhou2017activation}. 
Arguably, IS~\cite{salimans2016improved} is the most widely used metric for evaluating generative models. It employs a pre-trained neural network to evaluate the desired properties of generated samples, such as high classifiability and diversity of class labels. This metric shows a reasonable correlation with the quality and diversity of the generated images. However, IS is also known to have multiple limitations~\cite{yan2016attribute2image}. Firstly, IS is sensitive to over-fitting. Secondly, it may favor models that learn from diverse images. Also, operations such as mixing natural images from completely different distributions may cheat this metric. 

Another widely used evaluation metric for generative models is FID score~\cite{heusel2017gans}. To compute this metric, images are embedded into a feature space. Thereafter,  mean and covariance of both generated and real data embeddings are  computed and FID measures the image quality by comparing these two distribution parameters. While FID excels in discriminability, robustness, and speed, it assumes that the data features follow a Gaussian distribution, which is not always the case. It is also notable that  both IS and FID are model- or dataset-specific metrics. This makes them less suitable to quantify the quality of individual images.

\vspace{-2mm}
\section{Proposed Approach}
\vspace{-1mm}
Numerous works highlight the importance of texture, edges, and frequency for detecting fake images~\cite {do2005contourlet, chen2022deepfake,nguyen2019capsule, rossler2019faceforensics++, afchar2018mesonet, cozzolino2017recasting}. However, most fake detection methods proposed are still `learned' models. They face intrinsic limitations of carefully handling training demands while falling short on generalizing to unseen generative techniques~\cite{pashine2021deep}. 
In this work, we develop a non-learning based metric called Image Realism Score (IRS) to quantify realism in generated images, as illustrated in Fig.~\ref{Frame}. This metric incorporates five image statistics, and leads to a convenient detection of synthetic images generated by the contemporary diffusion models as well as BigGAN. In this section, we first introduce the mathematical principles behind the employed statistics and discuss how they positively influence our metric. Subsequently, we introduce their infusion into IRS. Later, we discuss the use of the proposed IRS for fake  content detection and improvement of the popular Stable Diffusion Model (SDM)~\cite{rombach2022high} using the same metric.

\vspace{-2mm}
\subsection{Image Statistical Measures}
\label{sec:imageMeasures}
\vspace{-2mm}
Gray-Level Co-occurrence Matrix  (GLCM)~\cite{sebastian2012gray} is used in image processing to extract textual features of an image. One of the statistical measures that can be derived from a GLCM is ``Energy", also known as ``Uniformity" or ``Angular Second Moment". Given a GLCM  $P \in \mathbb R^{N \times N}$  (where \( N \) is the number of gray levels in the image), the Energy $E$ is defined as

\begin{equation}
\text{GLCM}_E = \sum_{i=0}^{N-1} \sum_{j=0}^{N-1} [P(i,j)]^2,
\vspace{-4mm}
\end{equation}

where \( P(i,j) \) represents the joint probability of occurrence of pixel pairs with intensity values \( i \) and \( j \) at a specified spatial relationship. A higher Energy (E) score suggests that there are fewer variations in intensity in an image, indicating more uniform or repetitive patterns. In the process of removing noise or irregularities while generating images, generative techniques may blur some texture details. This can leave their signature in the $\text{GLCM}_E$.

The second metric, {GLCM Contrast}, quantifies the difference or change between adjacent pixels present in the image. Such differences or changes are indicative of the texture contrast in the image. The Contrast $C$ is defined as
\begin{equation}
\text{GLCM}_C = \sum_{i=0}^{N-1} \sum_{j=0}^{N-1} (i-j)^2 P(i,j),
\vspace{-2mm}
\end{equation}
where \( P(i,j) \) represents the probability that a pixel with intensity \( i \) co-occurs with a neighboring pixel of intensity \( j \) in a specified spatial relationship. A higher value of Contrast implies more variations in intensity between a pixel and its neighboring pixels across the image. Smooth and blurred generated textures are expected to result in lower GLCM Contrast values for generated images.

{Canny Edge Density}~\cite{tahmid2017density} refers to the proportion of pixels in an image that are identified as edges using the Canny edge detection technique. Given the total number of pixels \( I \) and number of edge pixels \( E \)  detected by the Canny edge detector, the Canny Edge Density, \( D \), can be defined as 
\begin{equation}
\text{CED} = \frac{E}{I}.
\vspace{-2mm}
\end{equation}
Here, CED $\in [0,1]$ represents the proportion of the edge pixels in the image. 
Higher CED value indicates high edge density. A visually appealing, generated images can be expected to differ from natural images in terms of their edge pixel distribution. Hence, Canny Edge Density is another helpful statistic for quantifying realism in images.

Variance Blur Measure (VBM)~\cite{he2005laplacian} is used to estimate the sharpness or blurriness of an image. The process involves computing the variance of an image after applying a Laplacian filter. For an image with dimensions \( M \times N \), the VBM is measured as, 
\begin{equation}
\text{VBM} = \frac{1}{M \times N} \sum_{i=1}^{M} \sum_{j=1}^{N} (L(i, j) - \mu)^2,
\vspace{-2mm}
\end{equation}
where \( L(i, j) \) is the pixel value at position \( (i, j) \) in the Laplacian filtered image and \( \mu \) represents the mean pixel value of the Laplacian filtered image. Due to the differences in smoothness between real and fake images, VBM is expected to generate different score in fake images.

The Mean Spectrum (MS)~\cite{frank2020leveraging} is a concept used in the frequency domain analysis of images. When an image undergoes a Fourier Transform, it produces a spectrum with both magnitude and phase components for each frequency. The MS provides an average measure of the spectrum magnitude. For an image with dimensions \( M \times N \) and its Fourier Transform  \( F(u,v) \), the Mean Spectrum is given as
\begin{equation}
\text{MS} = \frac{1}{M \times N} \sum_{u=0}^{M-1} \sum_{v=0}^{N-1} |F(u,v)|.
\end{equation}
Here, \( |F(u,v)| \) represents the magnitude at frequency coordinates \( (u, v) \). Image generation usually involves blending faces or objects from various sources, often introducing subtle artifacts~\cite{xiao2018generating}. By leveraging the Mean Spectrum, these inconsistencies can be brought to light.


\vspace{-2mm}
\subsection{Image Realism Score}
\vspace{-2mm}
Real images exhibit natural attributes, which directly result from natural scenes. On the contrary, attributes of generated images are dependent on the processes underlying the generative model. This can lead to unnatural image statistics for the generated images. The key intuition behind our Image Realism Score (IRS) is to scrutinize the primitive image attributes to quantify the realism of the image content. We do so by combining the above-mentioned five image statistics in our IRS.
Our technique enables leveraging the numerical differences in the values of the image statistics to distinguish between real and fake images.
\vspace{-2mm}
\subsubsection{Sort Order of the Measures}
\vspace{-2mm}
The five measures  we opt (refer Section~\ref{sec:imageMeasures}) to define IRS capture largely unrelated primitive statistics of images. To support this argument, we report the correlations between the five measures of ten thousand random images from the ImageNet dataset~\cite{russakovsky2015imagenet} in Table~\ref{tab:order}. Since the chosen measures eventually just provide numerical values, simply combining them using basic arithmetic operations is ineffective for our ultimate objective. Hence, we devise a unique strategy to maximize the collective information we can extract from our measures.
The computation method of our IRS requires defining a graph using the measures. For $n = 5$ measures that have equal importance for IRS, we choose pentagon as the base graph geometry. Each edge of the graph, which translates to the radii of the pentagon, signifies one of the used measures, see Fig.~\ref{Frame}.


\begin{tiny}
\begin{table}[t]
\scriptsize
\centering
\renewcommand\arraystretch{1.1}
\setlength{\tabcolsep}{2.5mm}{
\caption{Correlation coefficient between five image statistical measures. CED, GLCM$_C$, GLCM$_E$, VBM, and MS represent Canny Edge Density, GLCM Cantrast, GLCM Energy, Variance Blur Measure, and Mean Spectrum, respectively.}
\label{tab:order}
\vspace{-3mm}
\begin{tabular}{lccccc}
\hline
& CED  & GLCM$_C$ & GLCM$_E$ & VBM & MS \\
\hline
CED  & - & 0.76 & -0.28 & 0.05 & 0.21 \\
GLCM$_C$ & 0.76 & - & -0.13 & 0.20 & 0.14 \\
GLCM$_E$& -0.28 & -0.13 & - & 0.21 & -0.35 \\
VBM & 0.05 & 0.20 & 0.21 & - & -0.07 \\
MS & 0.21 & 0.14 & -0.35 & -0.07 & - \\
\hline
\end{tabular}}
\end{table}
\end{tiny}

The radii can be used to construct five triangles comprising the pentagon. Eventually, we use the areas of these triangles to compute the area of the pentagon that defines IRS. 
For our geometric graph shape, a triangle has the central angle \( \theta \).  
The area of the triangle delineated by two radii, say $m_a$ and $m_b$, can be computed as
\begin{equation}
A_{\triangle_{a,b}} = \frac{m_a \cdot  m_b}{2}  \cdot \sin\left(\theta\right).
\vspace{-2mm}
\end{equation}
Denote the radii of the pentagon as $m_1, m_2,...,m_5$. Since each radii signifies a unique measure, their order  matters for the ultimate IRS value. The order of the radii can vary. 
Initial combinations of two radii that are adjacent to, say $m_1$ can be calculated as
\begin{equation}
C = \frac{(n-1)!}{x!(n-x-1)!},
\vspace{-2mm}
\end{equation}
where $x = 2$ is the number adjacent radii and $n = 5$ are the total radii.
This leads to $C = 6$ combinations. 
For these combinations, the adjacent radii can be arranged in $2$ ways. 
Hence, the total number of combinations is $6 \times 2 = 12$.

\begin{figure}[t]
  \centering
\includegraphics[width=1.0\linewidth]{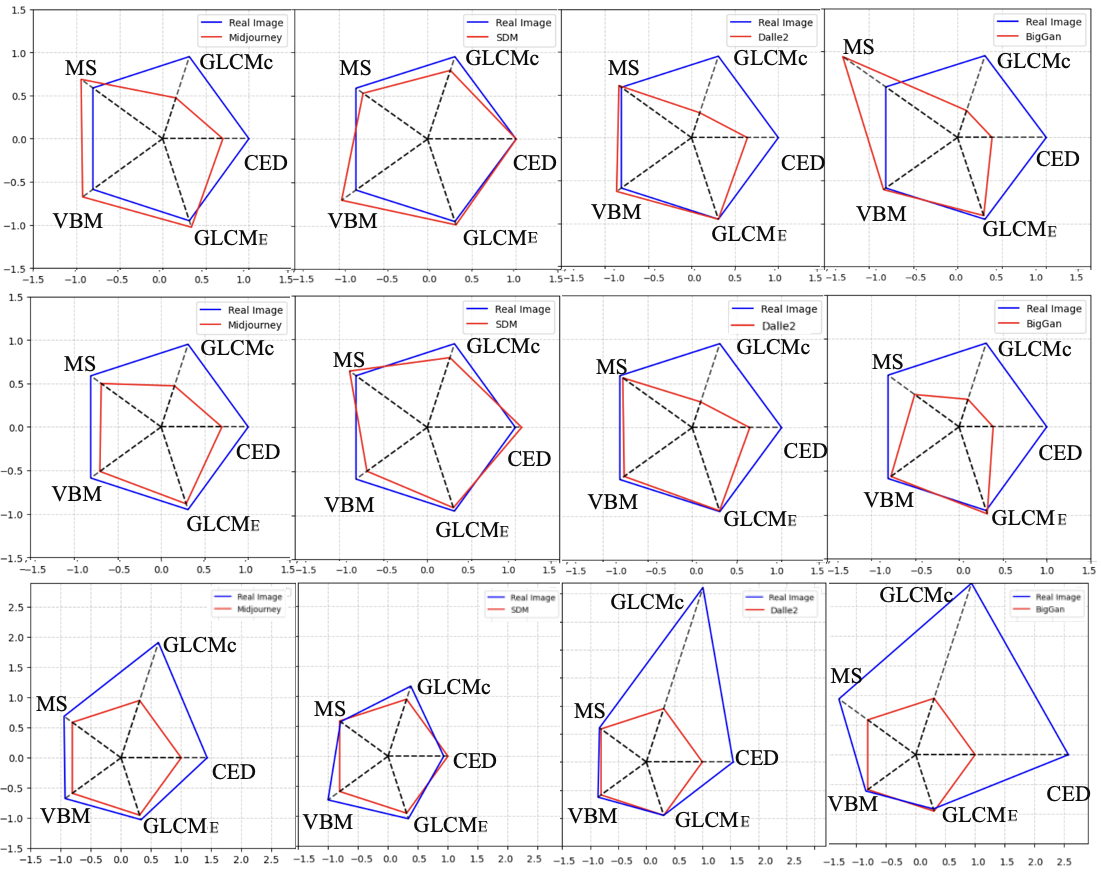}
\vspace{-5mm}
   \caption{In row 1 and row 2, the blue normalized pentagon is computed as the average of 10,000 ImageNet (real) images. Red pentagons are formed by the average of 1,000 images generated by the four models. In row 2, the MS, VBM, and GLCM$_E$ values are inverted for generated images. In row 3, the red and blue pentagons are re-scaled by the same ratio so that the red pentagons become uniform, which makes the blue pentagons nonuniform.}
   \label{reciprocal}
\end{figure}

To find a suitable order from the 12 combinations of metrics in the pentagon, we leverage their mutual correlations. 
Based on the empirical results in Table~\ref{tab:order}, we arrange the two measures with the largest correlation as adjacent radii and continue in a descending order.  
The intuition here is that this systematic  arrangement can lead to larger areas, which are eventually desirable for IRS for its discriminative abilities. 
To validate this intuition, we conducted experiments on 10,000 random ImageNet images, calculating and ranking areas for all 12 arrangements. We recorded that this arrangement resulted in the maximum area $\sim37\%$ times whereas the remaining 63\% was distributed between the other 11 arrangements. Finally, we express the area of a pentagon formed by a fixed sequence of radii as
\begin{equation}
A = \sum_{(a,b) \in S} A_{\triangle_{a,b}},
\vspace{-2mm}
\end{equation}
where $S \in \{ (1,2), (1,5), (2,3), (3,4), (4,5)\}$.

\subsubsection{Calibration of the Statistical Measures}
We have carefully chosen the  measures to be incorporated in our metric. However, each measure has its own numerical variability. This calls for a  calibration before combining them into IRS. In Fig.~\ref{reciprocal}(row-1), we show (red) pentagons of fake images that get formed for the four generative models used. We also provide a (blue) reference pentagon in each sub-figure that corresponds to the normalized values of the five measures resulting from real images. It can be noticed that the radii GLCM$_E$, VBN and MS of the (red) generated image pentagons are often exceeding the radii of the reference (blue) pentagon. The reason behind their behavior is as follows. 

It is known that generative techniques do not capture all subtle textural details  found in real images~\cite{arora2017gans}. This results in relatively smoother images which in turn leads to higher GLCM$_E$, as well as higher VBM. Similarly, image generation can  lead to inconsistencies in noise distribution, which get more pronounced in the frequency domain~\cite{mahdian2009using}. This causes relatively higher MS values for the generated images. The values of the other two measure are  loosely upper-bounded by their real image values. Hence, we take multiplicative inverses of VBM, MS and GLCM$_E$. This leads to polygons that are confined much better within the real image pentagons, see  Fig.~\ref{reciprocal}(row-2). Before further processing, we normalize the (red) pentagons of fake images and scale the (blue) pentagons of real images by the same proportion as shown in Fig.~\ref{reciprocal}(row-3). These two steps are helpful because our technique uses polygon area comparison for fake detection is Section~\ref{sec:FakeDetect}.

In Table~\ref{tab:weights}, we report the average values of the measures `before' and `after' their calibration. More precisely, the `before' calibration case corresponds to 
row-2 of Fig.~\ref{reciprocal}, which already accounts for VBM, MS and GLCM$_E$ variability. However, notice that the difference between the average areas of real and fake images is still  only 0.88. To amplify this difference, we re-scale the measures for fake images to 1.0, and also  re-scale the measures of  real images correspondingly (see Fig.~\ref{reciprocal}(row-3)) which increases the difference between the average areas to 2.30.  After this re-calibration, the eventual IRS value is computed as 
\begin{equation}
\label{Area}
\begin{split}
{\rm IRS}  = \sum_{(a,b) \in S} w_a \cdot w_b \cdot A_{\triangle_{a,b}}, 
\vspace{-4mm}
\end{split}
\end{equation}
where $w_a$ and $w_b$ are the weights resulting from the re-scaling of the corresponding measures.



\begin{tiny}
\begin{table}[t]
\scriptsize
\centering
\renewcommand\arraystretch{1.1}
\setlength{\tabcolsep}{1.2mm}{
\caption{Calibration of the statistical measures. Average values are reported for all. After calibration, the difference between the average IRS of real and fake images increases considerably.}
\label{tab:weights}
\vspace{-3mm}
\begin{tabular}{lccccr|r}
\hline

Metrics & GLCM$_C$  & GLCM$_E$   &  CID  & VBM  & MS
 & IRS\\
\hline
\hline
& & \textbf{Before}& \textbf{Calibration} &  \textbf{}\\
\hline
Real Images & 1.00 & 1.00 & 1.00& 1.00 & 1.00&2.38\\
Fake Images & 0.43
 & 0.97
 & 0.64
& 0.90
 & 0.98
 & 1.50
\\
\hline
& & \textbf{After}& \textbf{Calibration} &  \textbf{}\\

\hline
Real Images & 2.31 & 1.02 & 1.57& 1.11 & 1.02 &4.68\\
Fake Images & 1.00
 & 1.00
 & 1.00
& 1.00
 & 1.00
 &2.38\\

\hline
\end{tabular}}
\end{table}
\end{tiny}

\subsection{Fake Detection}
\label{sec:FakeDetect}
Our proposed metric can be evaluated on a single image, making it particularly suitable for detecting fake content on a sample-by-sample basis. This type of detection is not possible using traditional metrics such as IS and FID, which typically rely on large datasets. For fake content detection, the IRS adopts a thresholding technique. The last column in Table~\ref{tab:weights} shows that the average areas of real and fake images generally vary greatly under our proposed scheme. We capitalize on this observation, and use a threshold $\delta: \text{IRS}<\delta \implies \text{Fake}$, where $\delta = 3$ is set empirically in our experiments. This simple yet effective approach ensures efficient  fake image detection without the need for complex calculations.

\subsection{Gen-100: Dataset of Generated Images}
\label{sec:Dataset}

Another important contribution of this paper is the creation of an  new dataset, Gen-100. In addition to being used to evaluate the effectiveness of our method, this dataset can also be used for benchmarking future evaluation metrics. The Gen-100 data is generated using several popular image generation models, including SDM, BigGAN, Dalle2, and Midjourney. It consists of 100 object categories, where all categories follow CIFAR100~\cite{krizhevsky2012imagenet} categories. We use the aforementioned models to generate 1,000 images for each category. The real counterparts of the synthetic images are extracted from the same class labels of ImageNet~\cite{russakovsky2015imagenet}. We use  ChatGPT~\cite{openai2023gpt4} to generate 10 prompts for each category, which are used to further generate the text-conditioned images from the models.  This process allows us to guarantee  diversity in the dataset, while also capturing advanced abilities of the models. 
Due to the prevalent trend of non-open-source diffusion-based generative models that require payment for access, a void exists in the realm of comparative datasets comprising images generated by different diffusion models. Our dataset fills this gap. The dataset will be made public after acceptance.

\subsection{Improving Image Generation}
\label{sec:SDM}
Diffusion models (DMs)~\cite{sohl2015deep} learn data distributions through a unique approach. The core idea is to learn the distribution \( p(x) \) by consecutively denoising a variable that follows a normal distribution. This progressive denoising can be conceived as backtracking a Markov Chain of fixed length \( T \). An objective function is defined to gauge the denoising process as follows
\begin{equation}
L_{DM} = \mathbb{E}_{\epsilon(x), t,\epsilon}\left[||\epsilon-\epsilon_\theta(x_t,t)||^2\right],
\end{equation}
where \( t \) is a time step, uniformly chosen from the range \{1, \ldots, T\}, \( x_t \) represents the noisy version of the input at time \( t \), \( \epsilon_\theta(x_t,t) \) signifies the model's denoised prediction at time \( t \) and $L_{DM}$ is the model loss.
Following DMs, a Latent Diffusion Model (LDM) \cite{rombach2022high} generates images by iterating  over denoised data in a latent representation space, and then decode the representation results into full images. Its loss objective is defined as
\begin{equation}
L_{LDM} = \mathbb{E}_{\epsilon(x), t,\epsilon}\left[||\epsilon-\epsilon_\theta(z_t,t)||^2\right],
\end{equation}
were, \( z_t \) is the representation in the latent space, abstracting the input's finer details.

The widely popular Stable Diffusion Model~\cite{rombach2022high} is an LDM. To improve the realism in its generated images, we incorporate our IRS metric into the models training loss objective. Specifically, we minimize $\frac{1}{\text{IRS(image)}}$ because real images have large IRS values, see Table~\ref{tab:weights}. This encourages the model to generate more realistic images by minimizing the training loss. It is notable that we are able to compute IRS values on per-image basis here, which allows us to easily modify  the training objective of SDM. The improved training objective of the model is defined as 
\begin{equation}
L_{IRS} = \mathbb{E}_{\epsilon(x), t,\epsilon}\left[||\epsilon-\epsilon_\theta(z_t,t)||^2\right] + \frac{\xi}{{\rm IRS}(d(z_t))}, 
\end{equation}
were, $d$ is the decoding function of the original SDM approach, and $\xi$ is a scaling factor for our regularization. 


\begin{tiny}
\begin{table}[t]
\scriptsize
\centering
\renewcommand\arraystretch{1.5}
\setlength{\tabcolsep}{2.5mm}{
\caption{Average IRS values of  different generative models computed on 1K images from Gen-100 dataset. The value for Real images is computed for 10K ImageNet samples.}
\label{tab:Modelscampare}
\vspace{-3mm}
\begin{tabular}{lcccrcc}
\hline
Model Name
 & SDM  & Dalle2 
 &  Midjourney  
 & BigGAN & Real \\
 \hline
IRS score & 2.29 & 1.58 & 2.03& 1.74 & 4.68 \\
\hline
\end{tabular}}
\end{table}
\end{tiny}


\section{Experiments}

\begin{figure}[t]
  \centering
\includegraphics[width=1.0\linewidth]{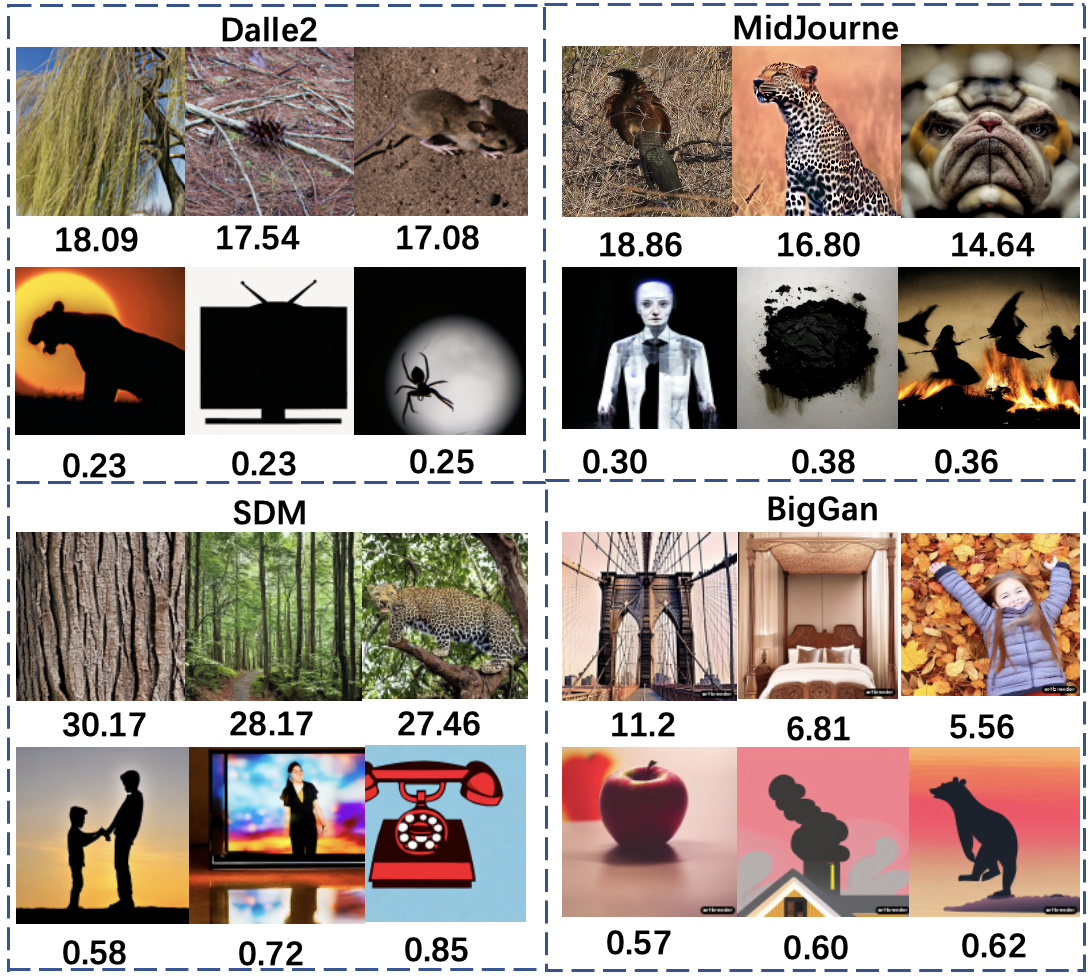}
\vspace{-7mm}
   \caption{Representative generated images for different generative models that led to high and low IRS values. Images with higher IRS are clearly more realistic.}
   \label{ModelImages}
   \vspace{-2mm}
\end{figure}

\noindent{\bf Benchmarking generative models: } We first benchmark the popular generative models using our metric. In the experiment, we employ our Gen-100 dataset - see Sec.~\ref{sec:Dataset}.
The average IRS values for the SDM, Dalle2, Midjourney and BigGAN models are reported in Table~\ref{tab:Modelscampare}. The table also includes the average value of 10K real images as a reference. It can be seen that the IRS values are generally in accordance with the known abilities of the models. Interestingly, whereas Midjourney is popular for its high quality images, it scores lower than SDM on our metric. This is because despite their high quality, Midjournay images lack in `realism' as compared to  SDM, and IRS is intended to quantify realism.
In Fig.~\ref{ModelImages}, we show representative images generated by each model for high and low IRS values. It is easily noticeable that images with high IRS values indeed contain details that make them appear highly realistic. On the other hand, images with low IRS values are indeed of cartoonic nature, lacking in realism. 


\vspace{1mm}
\noindent{\bf Fake Detection:}
In Table~\ref{tab:ACC}, we report the results of using IRS for fake image detection. For this experiment, we use random 500 samples for each model from the Gen-100 dataset and following Sec.~\ref{sec:FakeDetect}, we use $\delta = 3.0$ as the threshold value. The results are reported with the standard metrics of accuracy, F1 score, recall and precision. It is noticeable that the results are generally in accordance with the realism quality reported in Table~\ref{tab:Modelscampare}. That is, the model with the highest IRS score from Table~\ref{tab:Modelscampare}, i.e., SDM, has the lowest detection rate. However, Dalle2 is still able to maintain a lower detection rate than BigGAN despite its smaller IRS value in Table~\ref{tab:Modelscampare}. We find that this is due to the larger versatility of Dalle2 images, which allows samples to score relatively high IRS with more frequency, instead of scoring very large IRS on a few images to achieve higher average IRS. This observation is inline with the general understanding that diffusion based models are more versatile than GANs.   

\begin{tiny}
\begin{table}[t]
\scriptsize
\centering
\renewcommand\arraystretch{1.1}
\setlength{\tabcolsep}{2mm}{
\caption{Fake detection results. The dataset for each model consists of 500 Gen-100 samples and 500 real images from ImageNet.}
\vspace{-3mm}
\label{tab:ACC}
\begin{tabular}{lccccrr}
\hline
Dataset Name 
  & BigGAN 
 &  SDM  
 & Dalle2  &Midjourney \\

 \hline

Accuracy & 0.85  & 0.76 & 0.81 & 0.79 & \\
F1 Score &  0.87 & 0.68 & 0.79 &  0.77& \\
Recall & 0.95  & 0.71 & 0.77 & 0.81 & \\
Precision & 0.81  & 0.73 & 0.81 & 0.78 & \\

\hline
\end{tabular}}
\vspace{-4mm}
\end{table}
\end{tiny}

\begin{figure*}[t]
  \centering
  \includegraphics[width=0.93\textwidth]{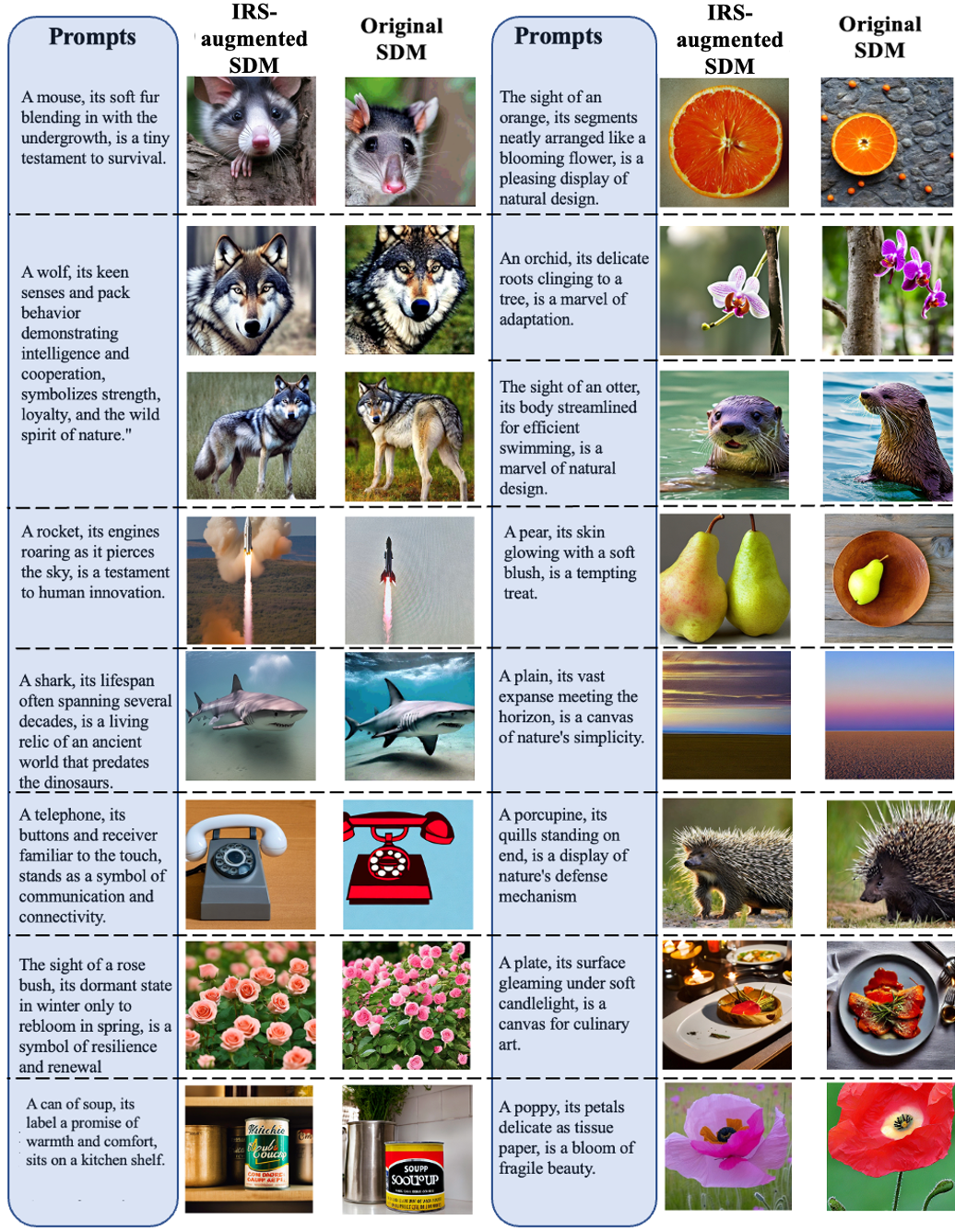}
    \vspace{-3mm}
   \caption{Comparison of generated images. One set is produced with the IRS-augmented SDM, and the other generated by the original SDM. Prompts used to generate the images are also given. Images generated by the proposed modified loss  showcase richer details, aligning with the essence of our technique.}
   \label{results1}
\end{figure*}

\vspace{1mm}
\noindent{\bf Improving SDM :}
We improved SDM with our technique following Sec.~\ref{sec:SDM}, and compare results with the original SDM. 
For a fair comparison, we use the exact same configuration for both models.
As shown in  Fig.~\ref{results1}, the IRS-augmented SDM produces images with sharper edges and clearer separation from the background. This makes the main subject in the image easier to identify. Secondly, the color transition is more natural, without abrupt color blocks or obvious artificial traces. In addition, the shadow and lighting effects presented by the improved model are more realistic. The images generated by the IRS-augmented SDM present finer textures, and both the subject and the background can clearly show their unique texture features. 

\section{Further Discussion}
\vspace{-1mm}

Currently, there are two popular metrics to measure the generative model quality, namely; Inception Score (IS)~\cite{salimans2016improved} and Fréchet Inception Distance (FID)~\cite{heusel2017gans}. However, both these metrics have their limitations~\cite{barratt2018note,chong2020effectively,obukhov2020quality}. 
For instance, the underlying classification model of IS adopts the Inception V3 architecture and is trained on the ImageNet dataset. Therefore, for unbiased evaluation using IS, generative models and classifiers are best trained on the ImageNet dataset. Additionally, sample size can also affect IS output. Moreover, insufficient samples may adversely affect IS results. Similar concerns are also valid for FID. In  contrast, our IRS eliminates the need for training, mitigating the risk of overfitting. Its design premise allows evaluation based on only a single image, avoiding sample size concerns.

In addition to the above-noted well-known shortcomings of IS and FID that have been verified in other papers, we also test  rotation invariance of these metrics. 
We performed random small rotations on the dataset and evaluated their impact on the IS and FID score. As shown in the Table~\ref{tab:rotation},  after rotation, the value of IS decreases and the value of FID increases. This means that for IS and FID, the same image can  become more ``fake" after rotation. In contrast, IRS uses only the most basic information of the image itself, which is not affected by rotations. This is an added advantage of our proposed metric.

\begin{tiny}
\begin{table}[t]
\scriptsize
\centering
\renewcommand\arraystretch{1.1}
\setlength{\tabcolsep}{2mm}{
\caption{Image rotation test with IS and FID. Each test dataset contains 1,000 generated images from the Gen-100 dataset.}
\label{tab:rotation}
\vspace{-3mm}
\begin{tabular}{lccccrr}
\hline
Dataset source
 & MidJourney  & BigGAN
 &  SDM  
 & Dalle2  \\

 \hline
 IS before rotation & 4.84 & 3.45 & 3.70 & 3.78 \\
IS after rotation& 4.35  & 3.15& 3.60 & 3.34 \\
\hline
FID before rotation & 3135.00 &2234.86  &2769.05  & 2550.84 \\
FID after rotation& 3389.25 &2700.14 & 2980.22 & 2784.39 \\
\hline
\end{tabular}}
\vspace{-4mm}
\end{table}
\end{tiny}

\vspace{-2mm}
\section{Conclusion}
\vspace{-1mm}
In recent years, the development of diffusion models has made it more difficult for humans to distinguish real and fake visual content. Therefore, exploring metrics for evaluating the authenticity of generated visual content is important. Our proposed Image Realism Score (IRS) addresses the shortcomings of existing metrics, such as the inability to analyze the authenticity of individual images and difficulties in achieving good results with tests on a new dataset different from the training sets. IRS avoids the limitations of current metrics by computing a fusion of five statistical measures of the input image. IRS is a non-learning based metric that does not rely on heavy computational resources. Using IRS, we also  successfully detected fake images generated by Stable Diffusion Model (SDM), Dalle2, Midjourney, and BigGAN (GAN), establishing the model-agnostic and data-independent nature of IRS. Furthermore, when IRS was incorporated in the loss function of SDM, the model performance was shown to improve.

\section{ACKNOWLEDGMENTS}
This research was supported by National Intelligence and Security Discovery Research Grants (project$\#$ NS220100007), funded by the Department of Defence Australia. Professor Ajmal Mian is the recipient of an Australian Research Council Future Fellowship Award (project number FT210100268) funded by the Australian Government.

{\small
\bibliographystyle{ieee_fullname}
\bibliography{egbib}
}

\end{document}